\documentclass[conference]{IEEEtran}
\IEEEoverridecommandlockouts
\usepackage{cite}
\usepackage{amsmath,amssymb,amsfonts}
\usepackage{algorithmic}
\usepackage{graphicx}
\usepackage{multirow}
\usepackage{glossaries}
\usepackage{textcomp}
\usepackage{xcolor}
\usepackage{makecell}
\usepackage{romannum}
\usepackage{tikz}
\usepackage{pgfplots}
\usepackage{url}
\usepackage{balance}

\def\BibTeX{{\rm B\kern-.05em{\sc i\kern-.025em b}\kern-.08em
    T\kern-.1667em\lower.7ex\hbox{E}\kern-.125emX}}

\usetikzlibrary{positioning, calc}

\definecolor{lavender}{rgb}{0.9, 0.9, 0.98}
\usepgfplotslibrary{colorbrewer}
\definecolor{c1}{HTML}{8DD3C7}
\definecolor{c2}{HTML}{FFFFB3}
\definecolor{c3}{HTML}{BEBADA}
\definecolor{c4}{HTML}{FB8072}
\definecolor{c5}{HTML}{80B1D3}
\definecolor{c6}{HTML}{FDB462}
\definecolor{c7}{HTML}{B3DE69}
\definecolor{c8}{HTML}{FCCDE5}
\pgfplotscreateplotcyclelist{spectralr-6}{%
Spectral-B,Spectral-E,Spectral-H,Spectral-J,Spectral-M,black!80}
\pgfplotscreateplotcyclelist{spectralr-3}{%
Spectral-B,Spectral-M,black!80}
\pgfplotscreateplotcyclelist{spectralr-1}{%
Spectral-M}
\pgfplotscreateplotcyclelist{iceoryx-2}{%
Spectral-M,Spectral-L}
\pgfplotsset{width=7cm,compat=1.18}

\begin{document}

\title{AutonomROS: A ReconROS-based Autonomous Driving Unit}

\author{
\author{\IEEEauthorblockN{
Christian Lienen\IEEEauthorrefmark{1},
Mathis Brede\IEEEauthorrefmark{2}, 
Daniel Karger\IEEEauthorrefmark{3}, 
Kevin Koch\IEEEauthorrefmark{4},
Dalisha Logan\IEEEauthorrefmark{5},\\
Janet Mazur\IEEEauthorrefmark{6},
Alexander Philipp Nowosad\IEEEauthorrefmark{7},
Alexander Schnelle\IEEEauthorrefmark{8},
Mohness Waizy\IEEEauthorrefmark{9},\
and Marco Platzner\IEEEauthorrefmark{10}}
\IEEEauthorblockA{\textit{Department of Computer Science} \\
\textit{Paderborn University}\\
\textit{Germany}\\
Email: \IEEEauthorrefmark{1}christian.lienen@upb.de,
\IEEEauthorrefmark{2}mbrede@mail.uni-paderborn.de,
\IEEEauthorrefmark{3}dkarger@mail.uni-paderborn.de,\\
\IEEEauthorrefmark{4}kevink2@mail.uni-paderborn.de,
\IEEEauthorrefmark{5}dalisha@mail.uni-paderborn.de,
\IEEEauthorrefmark{6}mazurj@campus.uni-paderborn.de,\\
\IEEEauthorrefmark{7}anowosad@mail.uni-paderborn.de,
\IEEEauthorrefmark{8}aschnell@mail.uni-paderborn.de,
\IEEEauthorrefmark{9}waizy@mail.uni-paderborn.de,
\IEEEauthorrefmark{10}platzner@upb.de}}

}

\maketitle

\begin{abstract}
Autonomous driving has become an important research area in recent years, and the corresponding system creates an enormous demand for computations. Heterogeneous computing platforms such as systems-on-chip that combine CPUs with reprogrammable hardware offer both computational performance and flexibility and are thus interesting targets for autonomous driving architectures. The de-facto software architecture standard in robotics, including autonomous driving systems, is ROS 2. ReconROS is a framework for creating robotics applications that extends ROS 2 with the possibility of mapping compute-intense functions to hardware. 

This paper presents AutonomROS, an autonomous driving unit based on the ReconROS framework. AutonomROS serves as a blueprint for a larger robotics application developed with ReconROS and demonstrates its suitability and extendability. The application integrates the ROS 2 package Navigation 2 with custom-developed software and hardware-accelerated functions for point cloud generation, obstacle detection, and lane detection. In addition, we detail a new communication middleware for shared memory communication between software and hardware functions. We evaluate AutonomROS and show the advantage of hardware acceleration and the new communication middleware for improving turnaround times, achievable frame rates, and, most importantly, reducing CPU load. 
\end{abstract}


\section{Introduction}
\label{sec:Introduction}

Autonomous driving is a rapidly evolving and important area of research~\cite{yurtsever2020survey}, which has the potential to fundamentally impact our transportation systems by, for example, significantly reducing traffic accidents~\cite{bertoncello2015ten, barabas2017current, yaqoob2019autonomous, parida2019autonomous}, congestion~\cite{yaqoob2019autonomous, parida2019autonomous, sun2020challenges}, and carbon dioxide emissions~\cite{yaqoob2019autonomous, montanaro2019towards, kopelias2020connected}. 

One of the major technical challenges that could impede the widespread adoption of autonomous driving is the high computational cost involved~\cite{zhao2020fusion, liu2019edge}. The computing demands for autonomous driving are exceptionally rigorous. Vehicles must continuously sense, process, and analyze their environment in real-time, make decisions, and execute them reliably and precisely.
Providing the required computational performance is expensive, and the resulting costs 
could make autonomous vehicles unaffordable for many people, especially as the technology is still in its early stages of development~\cite{yaqoob2019autonomous}. 
Additionally, the necessary electrical power needed for the computations can be substantial~\cite{liu2019edge}. This poses a particular problem for electric vehicles and can result in reduced ranges or larger and heavier batteries.
Moreover, the more complex autonomous driving systems become, the less reliable they become, and system failures may occur~\cite{chen2019model,lu2019cognitive}. 

One promising technology that addresses the computational cost challenge of autonomous driving is reconfigurable hardware. Reconfigurable hardware devices such as field-programmable gate arrays (FPGAs) comprise programmable logic and memory blocks and allow for exploiting parallelism in computations on several levels. Many modern FPGAs are systems-on-chip, including dedicated processors and even embedded GPUs. Compared to microprocessors, FPGAs exhibit massive parallelism, making them ideal for computationally intensive functions such as those seen in autonomous driving, e.g.,~\cite{li2022fpga, du2019fpga}, object detection~\cite{kojima2021autonomous, hao2019hybrid} or localization~\cite{liu2022energy}. Moreover, FPGAs can be highly energy-efficient~\cite{qasaimeh2019comparing, li2020ftrans, haripriya2022energy}. Compared to application-specific integrated circuits (ASICs), the key advantage of FPGAs is their flexibility, as they can be reprogrammed for different functions. 

Our work focuses on making hardware acceleration based on FPGAs available for robotics systems. Supporting widely used software architectures and programming abstractions is of utmost importance for the successful adoption of reconfigurable hardware technology. The standard software framework for modern robotics applications is ROS, the robot operating system, or ROS 2~\cite{ros2}. At the architecture level, ROS foresees the functional decomposition of an application into nodes, which are then linked via many-to-many publish-subscribe communication with ROS topics or one-to-one communication paradigms leveraged for ROS services and ROS actions. 

In the last years, several approaches for integrating reconfigurable hardware into ROS-based applications have been proposed~\cite{Sugata2017,8823798,mayoral2022robotcore,podlubne2020,reconfros}. One of these approaches is our own development ReconROS~\cite{LP20}, which allows for mapping complete ROS nodes to hardware while preserving the software programming model for both hardware and software nodes.

This paper presents AutonomROS\footnote{\url{https://github.com/Lien182/AutonomROS}}, an autonomous driving unit based on ReconROS. AutonomROS serves as a blueprint for a more extensive ReconROS-based application combining state-of-the-art open-source ROS 2 packages, custom-developed ROS 2 software nodes, and ROS 2 nodes completely mapped to hardware.
In particular, we make the following novel contributions:

\begin{itemize}
\item We present a new zero-copy communication middleware for ReconROS, based on Iceoryx, that greatly improves the performance of shared-memory inter-process communication between hardware and software nodes. 
\item We demonstrate hardware acceleration for the functions point cloud computation, obstacle detection, and adoption lane following. We show that AutonomROS is infeasible on the selected system-on-chip without hardware acceleration.
\item We show the suitability of ReconROS for developing larger robotics applications comprising existing software packages, ROS 2 software nodes, and nodes accelerated in hardware.
\end{itemize}

The remainder of the paper is structured as follows: Section~\ref{sec:BackgroundRelatedWork} provides background about ROS 2-based architectures for autonomous driving and ReconROS.  Section~\ref{sec:ReconROSExtensions} presents our new and efficient shared memory communication layer for ReconROS. In Section~\ref{sec:Architecture}, we elaborate on the architecture of AutonomROS and its hardware-accelerated components. Section~\ref{sec:Evaluation} presents an experimental evaluation, and Section~\ref{sec:Conclusion} concludes the paper and sketches ideas for feature work. 
\section{Background}
\label{sec:BackgroundRelatedWork}

In recent years, there has been much research in the field of architectures for autonomous driving. In this section, we first briefly introduce some relevant projects for autonomous driving and then provide background on ReconROS, the framework used by AutonomROS.

\subsection{Autonomous Driving Architectures}
\label{sec:BackgroundRelatedWork:HWAccelerationAutonomousDriving}

In 2015, the {\itshape Autoware Foundation} started the {\itshape Autoware.ai} project, which aimed to propose a functional software architecture for autonomous driving based on ROS 1 as an open-source project. Later, {\itshape Autoware.ai} was transferred to the {\itshape Autoware.auto} project~\cite{autoware}, now based on ROS 2, that provides a complete system stack for autonomous driving, including, e.g., functions for perception, planning, and control. Currently, {\itshape Autoware.auto} focuses on parking scenarios. The integration of FPGAs and the switch to a heterogeneous compute platform was shown in a follow-up project started by one of the {\itshape Autoware.ai} founders~\cite{chishiro2019towards}. Besides the architecture, the follow-up project also includes the specification and development of compilers, the operating system, middleware, and applications.

In another work, Reke et al.~\cite{reke2020self} proposed a self-driving architecture on ROS 2 that focuses on safe and reliable real-time behavior while preserving the main advantages of ROS, e.g., standardized message formats and distributed architecture. However, the proposed architecture does not include hardware acceleration and requires a desktop-class CPU for execution.

Several FPGA-based architectures for autonomous driving were presented in, for example, 
\cite{nitta2018study,yamamoto2021development,tanaka2019development,kojima2021autonomous,jones2019autonomous,wu2019end}. All these architectures have in common that they are designed for a rather limited scenario and do not rely on ROS 2 and state-of-the-art packages for autonomous driving, such as Navigation 2.

Additionally, there is work dealing with implementing and optimizing individual components of an autonomous driving architecture. Examples are provided in~\cite{9299131_lanelinedetection,gopinathan2022implementation}. In~\cite{9299131_lanelinedetection}, the authors present a lane line detection component for autonomous driving running on an embedded GPU. After color space transformation and filtering, the street lane is approximated by polynomial regression. The work is limited to the vision system of the autonomous driving architecture. The authors of~\cite{gopinathan2022implementation} describe a hardware-accelerated implementation of a lane detection algorithm on a Zynq-7000 System-on-Chip. The algorithm relies on edge detection on the input camera image, followed by a Hough transformation. Again, this work only considers the lane detection component. 

\subsection{ReconROS}
\label{sec:BackgroundRelatedWork:ReconROS}

ReconROS~\cite{LP20,lienen2021design} is a framework for implementing ROS 2 applications on heterogeneous compute platforms comprising a multi-core CPU and a reconfigurable hardware fabric and accelerating robotics application components on the reconfigurable hardware. ReconROS combines the ReconOS reconfigurable hardware operating system~\cite{Luebbers_Platzner_2009} with ROS 2.

\begin{figure}[h]
    \centering
    \includegraphics[width=0.75\linewidth]{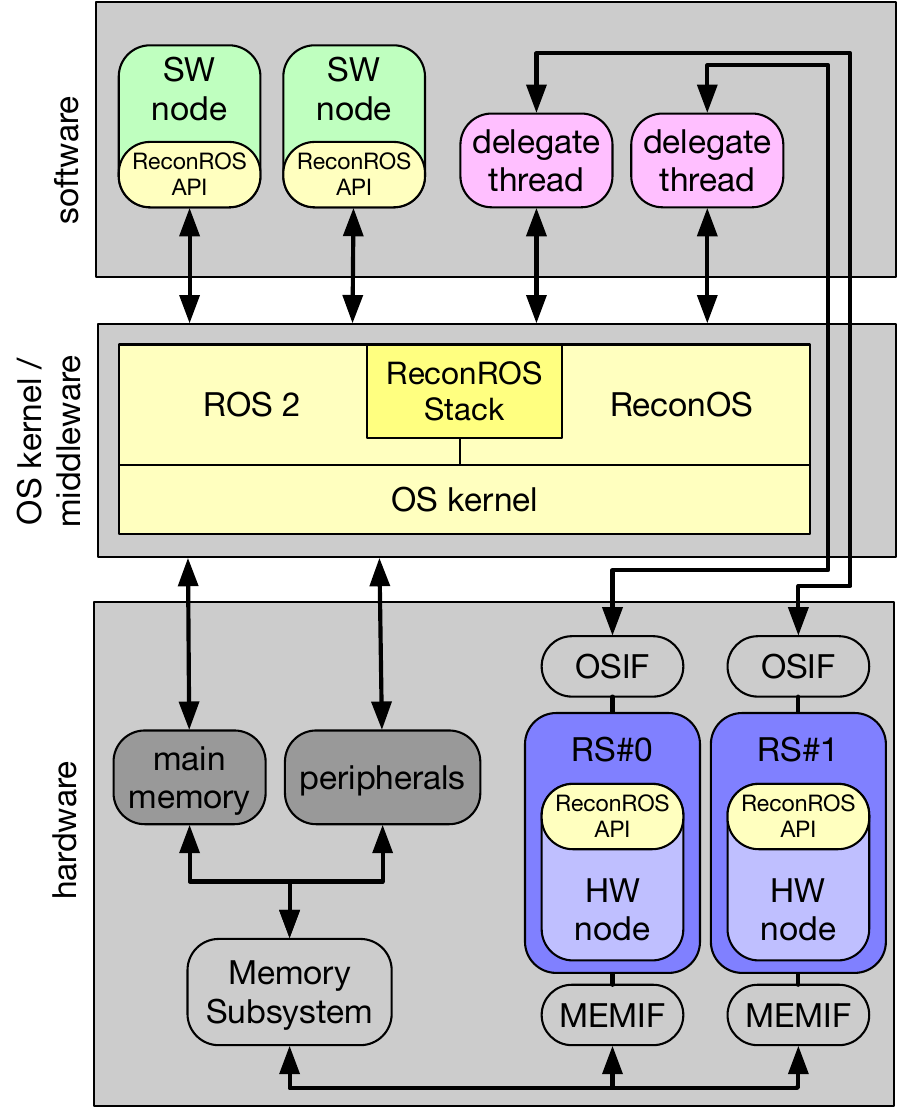}
    \caption{ReconROS Architecture (from~\cite{reconros_taskmapping}) }
    \label{fig:background:reconros_architecture}
\end{figure}

Figure~\ref{fig:background:reconros_architecture} sketches the hardware/software architecture of ReconROS. A ROS 2 application comprises a set of hardware nodes (HW node) and software nodes (SW node). HW nodes are assigned to specific rectangular regions on the reconfigurable fabric denoted as reconfigurable slots (RS). Each reconfigurable slot connects to the operating system kernel via an operating system interface (OSIF) and a lightweight software thread, the delegate thread. The delegate carries out operating system calls on behalf of the hardware node. Further, each reconfigurable slot connects to shared memory via a memory interface (MEMIF). This way, hardware nodes can access the shared virtual memory space, including any memory-mapped peripherals.

ReconROS extends the Linux operating system and enables the execution of ROS 2 nodes in hardware and software through the ReconROS stack and the ReconROS API. The ReconROS stack expands the capabilities of ReconOS by introducing ROS-related objects, such as ROS publishers or subscribers. Therefore, hardware and software nodes can interact with other ROS 2 nodes leveraging regular ROS 2 communication paradigms. Hence, communication is standardized, and integrating hardware-accelerated components into an existing ROS 2 application is greatly facilitated.

The ReconROS API is available for both software and hardware nodes and provides a consistent programming model across the software/hardware boundary. Hardware nodes can even be assigned to specific reconfigurable slots at runtime, allowing for dynamic mapping of a ROS application to nodes running in software and hardware.  

When a hardware node wants to leverage ROS 2 communication, it starts the interaction by sending a corresponding command via its OSIF to its delegate thread. Examples of such commands are a subscription to a ROS 2 topic or a request for a ROS 2 service. The delegate thread relays the command to the ReconROS stack and blocks until lower ROS 2 layers complete the processing of the requested command. When a message is received, the delegate is unblocked and transmits the location of the message in the main memory through the OSIF to the hardware node. The hardware node can then access the message through its MEMIF.
\section{ReconROS Shared-Memory Communication}
\label{sec:ReconROSExtensions}

A critical improvement of ROS 2 over ROS 1 was the introduction of an exchangeable communication layer based on well-established data distribution services (DDS). A developer can select between various available DDS implementations with different properties. Several DDS implementations rely on standard sockets as the default communication mechanism. Sockets are the most flexible option and enable both intra-platform and inter-platform communication. 
When intra-platform or even inter-process communication is required, a loopback adapter is employed to transfer data to the same or other processes. This flexibility is paid for with lowered performance since the loopback mechanism results in overheads due to several data copy operations involved. 

To mitigate such overheads, the Iceoryx~\cite{Iceoryx} communication middleware for ROS 2 was introduced. Iceoryx is an intra-process communication middleware enabling zero-copy data transmission between processes on the same platform. The disadvantage of Iceoryx, however, is that it comes with significant limitations, e.g., there is no support for ROS 2 services and actions. Since many larger software packages for ROS 2, e.g., Navigation 2, rely on these communication paradigms, the field of application for Iceoryx would be limited. Fortunately, Iceoryx is also part of the CycloneDDS middleware that allows for simultaneously using socket-based and shared-memory-based communication. When selecting shared memory for communication, the topic to which the ROS 2 nodes publish and subscribe has to adhere to the following constraints: (i) the message has a fixed length, (ii) a suitable QoS configuration is selected, (iii) the topic has at most 127 subscriptions, and (iv) a publisher has at most eight loaned messages simultaneously.

We have chosen the CycloneDDS middleware, with the integrated Iceoryx, as the basis for AutonomROS. Since Iceoryx requires a slightly different programming model than standard ROS 2 communication, we had to extend ReconROS to support the zero-copy communication scheme of Iceoryx. When using Iceoryx, a publishing node must first request a memory chunk from the middleware for communication with other nodes. The publishing node can write its message into the received memory chunk and execute the corresponding publishing function call if successful. Similar to standard ROS 2 subscribers, the subscribing node can block for a new message. However, after receiving it, the subscriber returns the message to the middleware to enable the re-usage of the message chunk. 

\begin{figure}[h]
    \centering
    \includegraphics[width=0.9\linewidth]{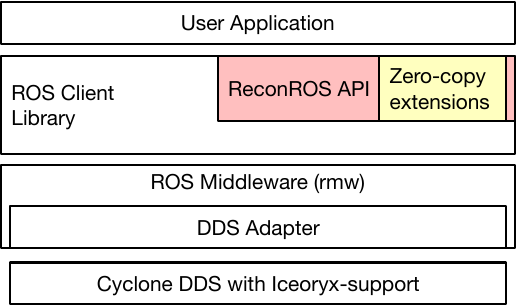}
    \caption{ReconROS API (red) extended by operations for zero-copy data transfers between nodes (yellow)}
    \label{fig:ReconROSExtensions:CommunicationStack}
\end{figure}

Figure~\ref{fig:ReconROSExtensions:CommunicationStack} shows the resulting extensions in the ReconROS API. Overall, we extended ReconROS by four function calls: 
{\ttfamily ROS\_BORROW} requests a memory chunk from Iceoryx, 
{\ttfamily ROS\_PUBLISH\_LOANED} publishes the message after the message has been written to the message chunk, 
{\ttfamily ROS\_SUBSCRIBE\_TAKE\_LOANED} tries to read a message, and 
{\ttfamily ROS\_SUBSCRIBE\_RETURN\_LOANED} returns the read message to Iceoryx so that the memory chunk can be re-used.

\section{Autonomous Driving Unit Architecture}
\label{sec:Architecture}

\begin{figure}[h]
    \centering
    \includegraphics[width=1.0\linewidth]{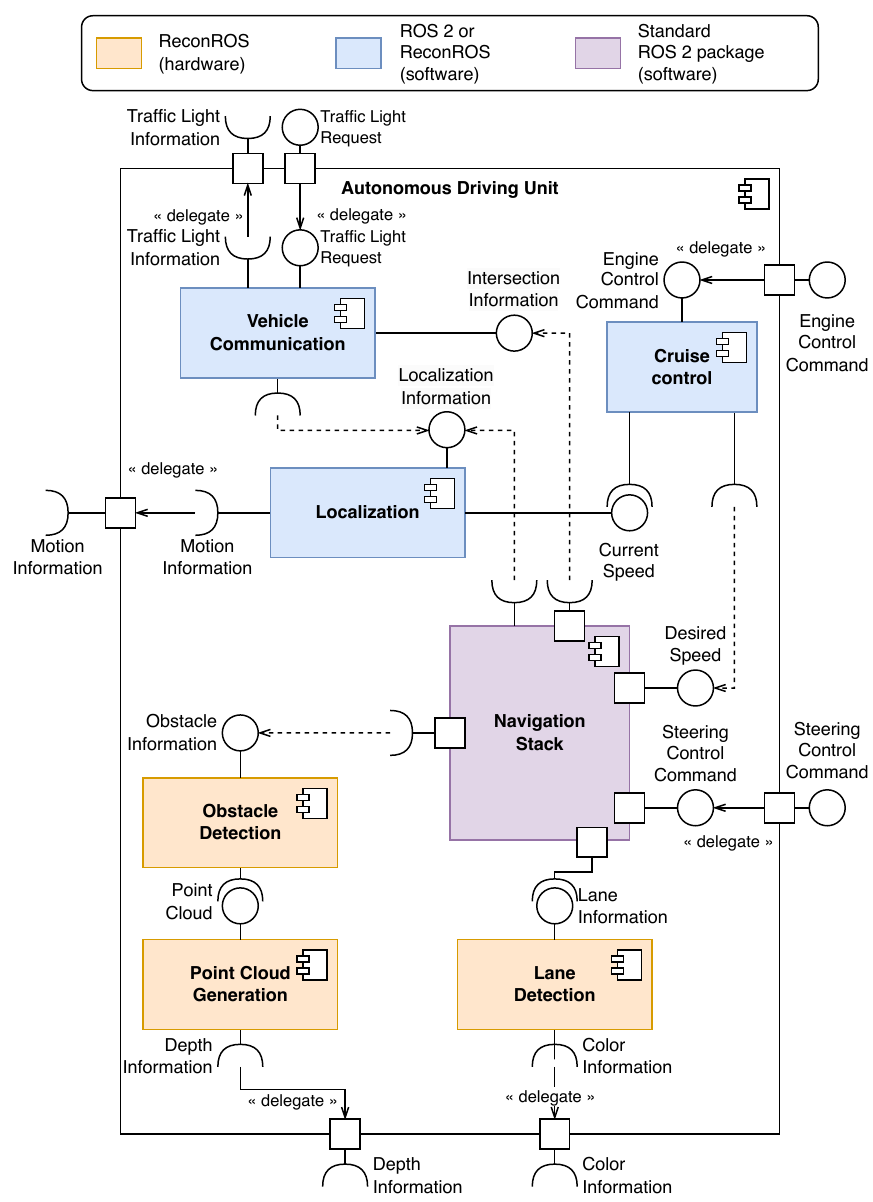}
    \caption{Architecture of the AutonomROS autonomous driving unit. Hardware-accelerated components are highlighted in blue color}
    \label{fig:Architecture:DrivingUnit}
\end{figure}

Figure~\ref{fig:Architecture:DrivingUnit} shows the top-level overview of the architecture of the AutonomROS autonomous driving unit. Besides providing its indented functionality, the architecture also aims to demonstrate that our ReconROS framework for creating robotics applications enables efficient hardware acceleration while maintaining the programming abstractions of ROS 2 and the ability to integrate larger ROS 2 software packages. 

AutonomROS comprises seven main components: The {\itshape Obstacle Detection} component and its preprocessing component {\itshape Point Cloud Generation} detect obstacles in front of the car, and the {\itshape Lane Detection} analyzes lanes. The {\itshape Navigation Stack} subsequently uses this information, which sets commands for steering control and the desired speed. The {\itshape Localization} component fuses different external sensors, e.g., inertial measurement units or wheel encoders, that provide information about the actual movement and the current position based on a static map. The {\itshape Vehicle Communication} component handles communication with the infrastructure around the car, e.g., a traffic light controller. It uses information about the vehicle's actual position from the {\itshape Localization} component to determine if the car is approaching an intersection. An entrance request is sent if the vehicle wants to enter an intersection. Eventually, the vehicle is allowed to enter the intersection. This permission is provided to the {\itshape Navigation Stack} component. The {\itshape Cruise Control} component controls the car's speed in a control loop leveraging a PID controller. The reference value of the control loop is the desired speed from the {\itshape Navigation Stack}. The difference between this reference value and the actual speed from the {\itshape Localization} component serves as the measured error for the PID controller. The output of the controller is forwarded to the engine of the vehicle.

Three of the components, {\itshape Point Cloud Generation}, {\itshape Obstacle Detection}, {\itshape Lane Detection}, show high computational demands with significant amounts of data processed and are thus suitable for hardware acceleration. We discuss these components in
more detail on the algorithmic level in Subsections~\ref{sec:Architecture:PointCloudGeneration}, \ref{sec:Architecture:ObstacleCollision}, and~\ref{sec:Architecture:LaneDetection}. These components are developed in C/C++ and can either be compiled using GCC for software execution, or synthesized with Xilinx Vitis HLS for hardware execution. Except for the communication with the traffic light, all communication is realized using standard ROS 2 publish-subscribe communication. The communication between the car and traffic lights relies on MQTT (Message Queuing Telemetry Transport).

Three more components, {\itshape Localization}, {\itshape Vehicle Communication}, and {\itshape Cruise Control}, are custom-designed for AutonomROS and mapped to ROS 2 software nodes. Finally, the component {\itshape Navigation Stack} is based on Nav2 (Navigation 2), an open-source ROS 2 package. We elaborate on this component in Subsection~\ref{sec:Architecture:NavigationStack}.

\begin{figure}[h]
    \centering
    \includegraphics[width=1.0\linewidth]{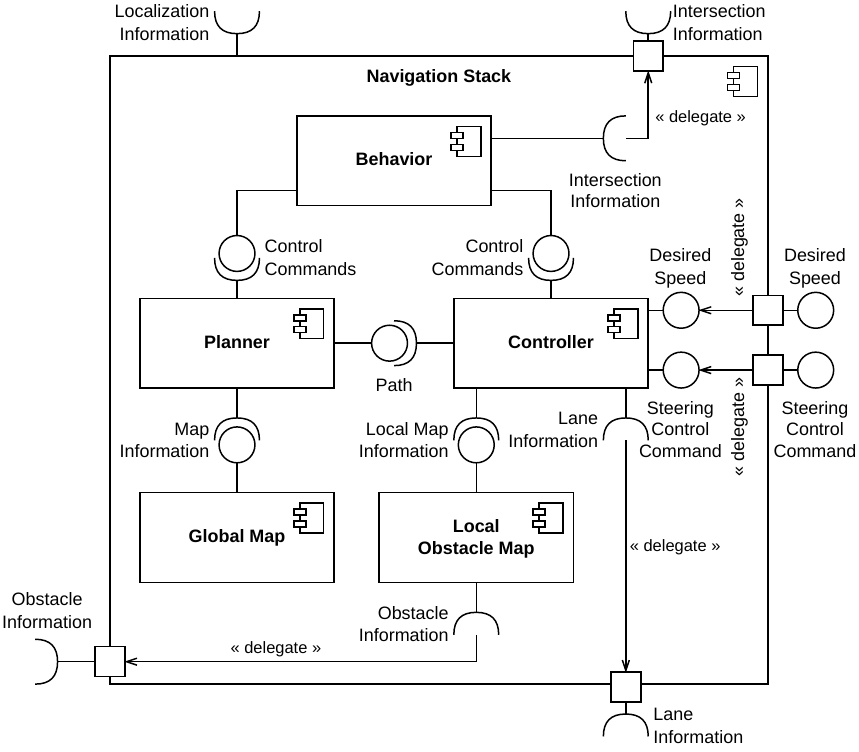}
    \caption{AutonomROS Navigation Stack}
    \label{fig:Architecture:NavigationStack}
\end{figure}

\subsection{Navigation Stack}
\label{sec:Architecture:NavigationStack}

Figure~\ref{fig:Architecture:NavigationStack} details the {\itshape Navigation Stack} component. It consists of the components {\itshape Behavior}, {\itshape Planner}, {\itshape Global Map}, {\itshape Controller}, and {\itshape Local Obstacle Map}. The whole component uses localization information for position estimation, which is crucial for all its sub-components. The {\itshape Behavior} component defines how the car should behave in different situations. It receives a goal position, which represents to which location the vehicle should drive. Additionally, it uses the intersection information to decide whether the car must stop in front of an intersection. These decisions result in different control commands for the {\itshape Planner} and {\itshape Controller}.
Among others, the {\itshape Planner} receives a goal position that the car should reach from 
the {\itshape Behavior} component. The {\itshape Planner} calculates, based on static map information from the {\itshape Global Map} and the car's current position, a path from the current position to the goal position. This path is then handed to the {\itshape Controller}. The {\itshape Controller} component also receives control commands from the {\itshape Behavior} component. These commands include whether the car should follow the path, lane, or stop because an intersection is blocked. The component calculates a steering control command and the desired speed based on the path or lane information. Additionally, the {\itshape Controller} uses information from a local map from the {\itshape Local Obstacle Map} to check whether an obstacle is in the way, and if so, stops the car.

\subsection{Point Cloud Generation}
\label{sec:Architecture:PointCloudGeneration}

The {\itshape Point Cloud Generation} component receives depth information from an external sensor, e.g., a 3D camera, and calculates a point cloud from the depth and corresponding color images. The {\itshape Obstacle Detection} component uses the resulting point cloud to detect obstacles in front of the car. Calculating a 3D point cloud involves analyzing and processing a depth image to generate a comprehensive representation of a physical object or environment in 3D. For point cloud computation, the first step is to merge the depth and the color image. Since pixels in the depth image are independent, this task can be easily parallelized and is ideally suited for hardware acceleration with ReconROS. 

Our hardware implementation of the component initially receives the camera's projection matrix $P$ as shown in Equation~\ref{eqn:hardware_acceleration:pointcloud:Pmatrix}. This 
matrix is needed to transform pixels from the depth image into the 3D world. It comprises the focal lengths $(fx,fy)$, the principal point $(cx,cy)$, and information about the relative position of the second camera to the first $(Tx, Ty)$ ~\cite{CameraInfo}. The matrix is published as a ROS 2 {\ttfamily CameraInfo} message by the camera's wrapper node. We need to gather this matrix only once before the actual runtime loop starts because the matrix does not change as long as the camera is not switched.

\begin{align}
    \label{eqn:hardware_acceleration:pointcloud:Pmatrix}
    P&=
    \begin{bmatrix}
        fx &  0 & cx & Tx \\
        0  & fy & cy & Ty \\
        0  &  0 &  1 &  0
    \end{bmatrix}
\end{align}

In the runtime loop, we transform all pixels of an incoming image with their coordinates $x$, $y$, and depth $w$ to their 3D world coordinates $X$, $Y$, and $Z$. To this end, we first determine intermediate variables $u=x\cdot w$ and $v=y\cdot w$ and then apply Equations~\ref{eqn:hardware_acceleration:pointcloud:eq3} - \ref{eqn:hardware_acceleration:pointcloud:eq5}, which are taken from the description of the {\ttfamily CameraInfo} message of ROS~\cite{CameraInfo}.

\begin{align}
    \label{eqn:hardware_acceleration:pointcloud:eq3}
    X &= \frac{u - cx \cdot w - Tx}{fx}\\
    \label{eqn:hardware_acceleration:pointcloud:eq4}
    Y &= \frac{v - cy \cdot w - Ty}{fy}\\
    \label{eqn:hardware_acceleration:pointcloud:eq5}
    Z &= w
\end{align}

\subsection{Obstacle Collision}
\label{sec:Architecture:ObstacleCollision}

Obstacle detection is typically done by computing a cost map layer based on the generated point cloud. Within the Navigation 2 package, a so-called Voxel Layer constitutes the default cost map layer. However, the Voxel Layer sequentially iterates over every point in the point cloud, which is slow. Thus, we have decided to replace the Voxel Layer by (i) processing the obstacle detection in hardware and (ii) handling the resulting data in the Navigation 2 package by a customized cost map layer.

The process of converting the point cloud into an obstacle grid is split into the following four steps: 
First, we transform the image from the camera's coordinate system into the car's base coordinate system based on a fixed transformation matrix. 
Second, we select all points in a predefined volume in front of the car and consider only the points in this "obstacle box" in the following steps. Since points higher than the car and points far away from the front or the sides of the vehicle need not be considered, this selection helps save on computations. 
Third, we project the points within the obstacle box to the ground, i.e., to the $xy$-plane.  
Finally, we discretize the obstacle box to a grid and assign to each grid cell the number of points of the original point cloud that map to the cell. The discretization reduces the required memory for representing obstacles from $4.7$ MB for the point cloud to $234$ Byte for the grid. In addition, the discretization reduces noise from the camera's depth sensor, which could result in false-positive detections of obstacles. 

Most of the involved processing is performed pixel-parallel. The final grid is published to a ROS 2 topic to make it usable for the custom cost map layer in the Navigation 2 package that runs in the software.

\subsection{Lane Detection}
\label{sec:Architecture:LaneDetection}

The {\itshape Lane Detection} component includes multiple computationally expensive image processing steps, e.g., the perspective and color thresholding transformation. Our component implementation involves several steps and relies on the open-source Vitis Image processing library\footnote{\url{https://github.com/Xilinx/Vitis_Libraries.git}}.

The first step transforms the incoming image from the RGB to the HSV color space. The transformation supports processing different color values independent of the environment's saturation or lighting.
The next step applies color thresholding to identify the image's white and yellow areas. 
Thresholding for both colors is processed in parallel, resulting in one grayscale image representing yellow and white pixels in the original image. 
The following step performs a warp transformation to the grayscale image based on a 
fixed transformation matrix to get a bird's view of the represented scene. After warp transformation, a decision is taken considering the number of pixels for each of the two colors, whether to follow the street's white or yellow lane. Subsequent processing focuses on the selected color.

Then, we perform a polynomial least-squares regression on the lane to eliminate interfering falsely detected pixels and establish an equation for the lane marking. Based on the $N$ pixels with coordinates $(x_n, y_n)$ that represent the lane and $k$ as the polynomial order to fit, we compute the desired polynomial coefficients $\vec{a} = [a_0 \dots a_k]^T$ by solving the equations system as shown in Equation~\ref{eq:hardware_acceleration:lane_detection:formula1}. 

\begin{equation}
    \begin{bmatrix} 
        N  & \dots &\sum_{k=0}^N x_{n}^k \\
        \vdots & \ddots & \vdots\\
        \sum_{n=0}^N x_{n}^k & \dots & \sum_{k=0}^N x_{n}^{2k} 
    \end{bmatrix}
    \cdot
    \vec{a}
    =
    \begin{bmatrix}
        \sum_{n=0}^N y_{n}\\
        \vdots\\
        \sum_{n=0}^N y_{n} x_{n}^k\\
    \end{bmatrix}
    \label{eq:hardware_acceleration:lane_detection:formula1}
\end{equation}

For implementing the {\itshape Lane Detection} component, we have determined that the second-order polynomial shown in Equation \ref{eq:hardware_acceleration:lane_detection:polynomial} is suitable. 

\begin{equation}
    f_l(x) = a_2 \cdot x^2 + a_{1} \cdot x + a_{0}
    \label{eq:hardware_acceleration:lane_detection:polynomial}
\end{equation}

To estimate the final trajectory, the resulting polynomial has to be shifted to the middle of the image and transformed into the car's base coordinate system.

For more efficient computation of the final trajectory, we use 30 equally-distributed points along the height of the image ($x_i = i \cdot 480/30$, $i = 0..29$) and compute 30 function values $y_i = f_l(x_i)$ (Equation \ref{eq:hardware_acceleration:lane_detection:polynomial}). 

Using the resulting 30 coordinates $(y_i,x_i)$, the equation system \ref{eq:hardware_acceleration:lane_detection:formula1} is solved again for a target polynomial function $f_t(x)$ (Equation \ref{eq:hardware_acceleration:lane_detection:polynomial_trajectory}).

\begin{equation}
    f_t(x) = a_3 \cdot x^3 + a_2 \cdot x^2 + a_{1} \cdot x + a_{0}
    \label{eq:hardware_acceleration:lane_detection:polynomial_trajectory}
\end{equation}

\section{Evaluation}
\label{sec:Evaluation}

In this section, we first report on the evaluation setup, including a real-world model car used for driving experiments to test the functionality of the AutonomROS driving unit. Then, we present architecture exploration experiments to evaluate the performance of different DDS versions and the hardware acceleration.

\subsection{Evaluation Setup}
\label{sec:Evaluation:Setup}

We execute AutonomROS on a Zynq UltraScale+ MPSoC ZCU104 evaluation board. The board contains a system-on-chip architecture with a quad-core ARM Cortex-A53, a dual-core Cortex-R5 real-time processor, a Mali-400 MP2 embedded graphics processing unit, and programmable logic (PL). For our evaluations, we used the quad-core CPU and the programmable logic. The board runs Ubuntu 20.04 in combination with ROS 2 galactic and ReconROS. 

\begin{figure}[h]
    \centering
    \includegraphics[width=1.0\linewidth]{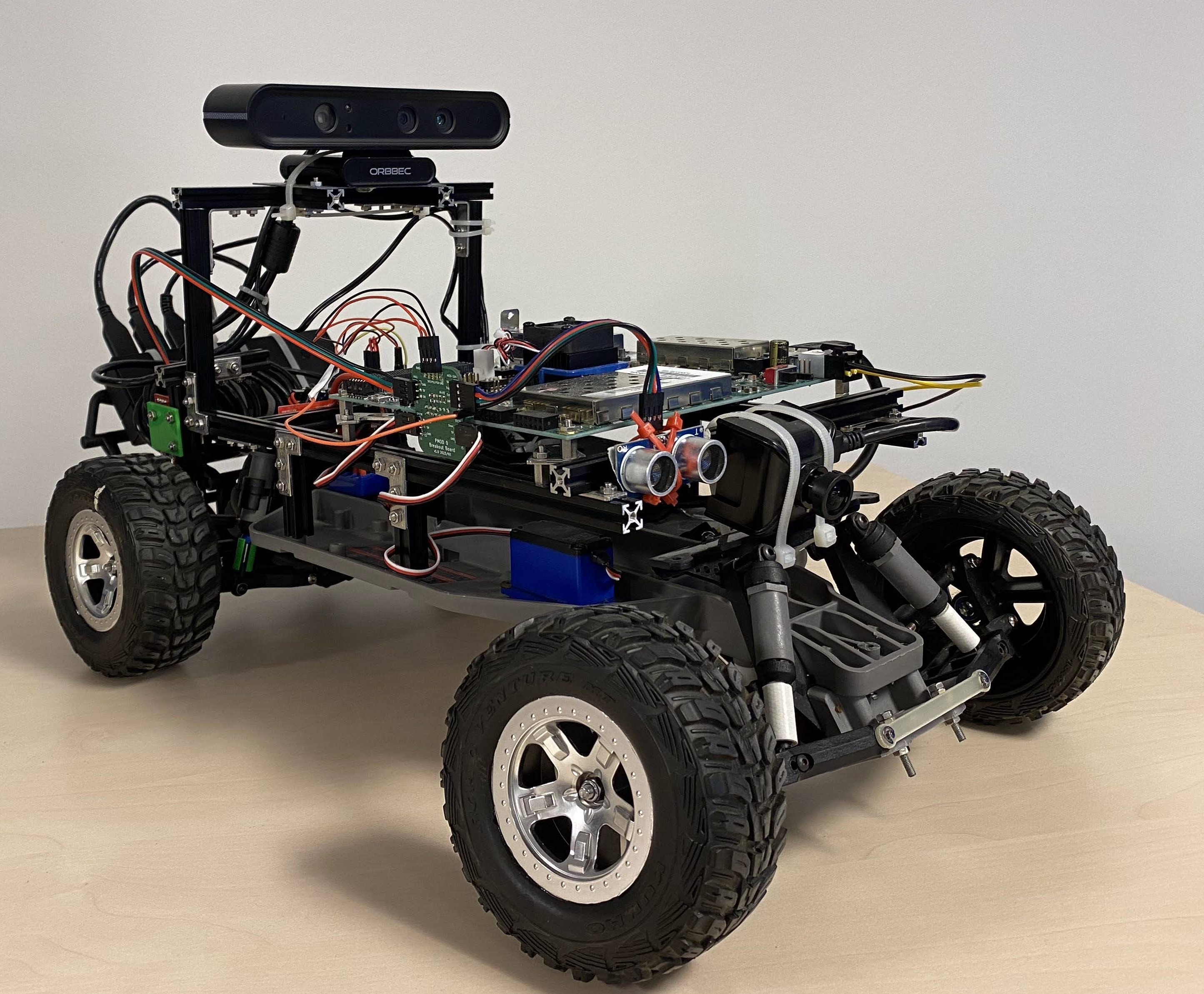}
    \caption{Model car platform}
    \label{fig:evaluation:Modelcar}
\end{figure}

We have mounted the evaluation board on a model car platform shown in Figure~\ref{fig:evaluation:Modelcar}. 
The model car platform is based on a modified commercial remote-controlled car in which the control and sensor systems have been replaced.
The actuators for the steering and the drive were preserved.
Regarding sensors, the platform includes two cameras, one for color information and one 3D camera providing depth and color information, an inertial measurement unit (IMU) for measuring acceleration data, and a wheel encoder for gaining speed data. Regarding actuators, the platform exhibits interfaces to drive the engine and the car's steering. Further, the evaluation board is equipped with a wireless LAN interface for data exchange with other vehicles and the infrastructure, e.g., the traffic lights controller. We have set up a $5m \times 5m$ grid of streets with two intersections to mimic real-world environmental test conditions. Additionally, we have set up a central infrastructure server that acts as a 
traffic lights controller and handles requests for crossing intersections. We have built two model cars to evaluate the AutonomROS functionality involving multiple vehicles.

\subsection{Performance Evaluation}
\label{sec:Evaluation:Performance}

Table~\ref{tab:measurements} summarizes the results of the performance measurements for the hardware-accelerated components {\itshape Point Cloud Generation}, {\itshape Obstacle Detection}, and {\itshape Lane Detection}. 
For these processing components, the test setup includes two ROS 2 nodes, one node publishing camera data, and one node including the actual processing. 
The table lists different architecture configurations, the total CPU load in \% of four cores, the achieved frames per second (FPS), and the turnaround time.
The turnaround time represents the node's raw computation time, which, in contrast to the FPS metric, is not limited by the input signal rate.

\begin{table}
    {
    \centering
    \renewcommand{\arraystretch}{1.2}
    \begin{tabular}{m{1.2cm}m{1.5cm}m{1.0cm}cc}
        Component & Architecture configuration & CPU$^{a}$ & FPS & \makecell{Turnaround\\time (ms)}\\
        \hline
        \hline
        \multirow{4}{*}{\rotatebox[origin=c]{0}{\makecell{Point Cloud \\ Generation\\@640x480\\ depth image}}}
        & \makecell{ReconROS SW\\without Iceoryx}   & 180--220  & 9--12   & 83--111\\
        \cline{2-5}
        & \makecell{ReconROS HW\\without Iceoryx}   & 180--210          & 13--14   & 71--76\\
        \cline{2-5}
        & \makecell{ReconROS SW\\with Iceoryx}      & 90--110           & 29--30   & 11--15 \\
        \cline{2-5}
        & \makecell{ReconROS HW\\with Iceoryx}      & 64--78            & 29--30   & 17--18 \\
        \hline \hline
        \multirow{2}{*}{\rotatebox[origin=c]{0}{\makecell{Obstacle\\Detection}}} 
        & \makecell{ReconROS SW\\ with Iceoryx}     & 50--60            & 29--30    & 15--17 \\
        \cline{2-5}
        & \makecell{ReconROS HW\\ with Iceoryx}     & 4--8  & 29--30    & 9--11 \\
        \hline \hline
        \multirow{2}{*}{\rotatebox[origin=c]{0}{\makecell{Lane\\Detection}}} 
        & \makecell{ROS 2 SW\\ with Iceoryx}           & 170--190          & 29--30    & 27--31\\
        \cline{2-5}
        & \makecell{ReconROS HW\\ with Iceoryx}     & 24--38            & 29--30    & 9--18 \\
        \hline
    \end{tabular}
    }\\
    $^{a}$CPU load in \% of 4 cores
    \caption{
    Performance measurements for the hardware-accelerated components of AutonomROS. The table shows min/max values collected in multiple measurements.
    }
    \label{tab:measurements}
\end{table}

\subsubsection*{Point Cloud Generation}
Two are two main observations: First, using the zero-copy Iceoryx middleware is highly beneficial, as expected, for both software and hardware mappings of the component. 
The CPU load and the achieved FPS are improved by roughly $2 \times $, and the turnaround time decreased by $7.5 \times$ compared to standard ROS 2 communication.
Adding hardware acceleration without a zero-copy communication middleware gives a minimal advantage. 
Second, using Iceoryx and hardware acceleration reduces the CPU load further, while the achieved FPS is bound by the camera's maximum frame rate of around 30 frames per second. 
The turnaround time is slightly higher because of overheads for data transmission into the programmable logic.

\subsubsection*{Obstacle Detection}
Here, we compare the software and hardware configurations with Iceoryx. 
Both configurations achieve the maximum FPS. 
For the hardware-mapped node, the CPU utilization of the component decreases significantly by a factor of $12.5 \times$, and the turnaround time is about $1.6 \times$ lower compared to the implementation in software. 

\subsubsection*{Lane Detection}
We compare a ROS 2 software implementation with a ReconROS hardware implementation. 
Compared to ReconROS SW implementation, this node relies on a standard ROS 2 C++ implementation, including the ROS executor enabling the event-driven programming model. 
Again, hardware acceleration results in a significant reduction of the CPU load and turnaround times.

The main conclusion of the measurements reported in Table~\ref{tab:measurements} is that for intra-platform communication, using a zero-copy communication middleware such as Iceoryx is of utmost importance to maintain performance. Moreover, effective hardware acceleration relies on such a middleware. The reported measurements are only for two ROS 2 nodes. Mapping the overall  AutonomROS unit of Figure~\ref{fig:Architecture:DrivingUnit} entirely to software would exceed the maximum CPU utilization of 400 \%. Thus, running the presented AutonomROS on the chosen system-on-chip platform is only possible with hardware acceleration.

\section{Conclusion and Future Work}
\label{sec:Conclusion}

In this paper, we have presented AutonomROS, a ReconROS-based unit for autonomous driving. AutonomROS integrates a ROS 2 open-source package for navigation with custom-developed software nodes for localization, cruise control, and vehicle communication and hardware-accelerated nodes for point cloud generation, obstacle detection, and lane detection. We have also introduced a new communication middleware for ReconROS that boosts the performance of shared memory communication between software and hardware nodes. Hardware acceleration combined with the new communication middleware results in significant performance improvements, measurable turnaround times, and achievable frame rates, but most pronounced in reduced CPU utilization with up to $12.5 \times$ for selected ROS 2 nodes.

In feature work, we plan to extend the AutonomROS driving unit with more advanced functionality, such as ORB-SLAM and GPS for localization or Advanced Driver Assistance Systems (ADAS). The advanced functionality should increase the potential for hardware acceleration.

\balance
\bibliographystyle{IEEEtran}
\bibliography{lienen22_irc}
\end{document}